\documentclass[10pt, a4paper]{article}

\usepackage[final]{lrec2026} 

\usepackage{makecell}
\usepackage{tablefootnote}
\usepackage{verbatim}
\usepackage{multirow}
\usepackage[flushleft]{threeparttable}
\usepackage[table]{xcolor}
\usepackage{booktabs}

\newcommand\modelformat[1]{\textsc{#1}}
\newcommand\mage{\modelformat{mage}}
\newcommand\mageood{\modelformat{mage-ood}}
\newcommand\mageoodpara{\modelformat{mage-ood-para}}

\newcommand\raid{\modelformat{raid}}

\newcommand\hcsi{\modelformat{h3c+ si}}
\newcommand\hcqa{\modelformat{h3c+ qa}}

\newcommand\mgt{\modelformat{m4gt}}

\newcommand\all{\modelformat{all}}

\title{Spotlights and Blindspots: \\ Evaluating Machine-Generated Text Detection}

\name{Kevin Stowe, Kailash Patil} 

\address{Pindrop \\
        \{kevin.stowe, kpatil\}@pindrop.com\\}

\abstract{
With the rise of generative language models, machine-generated text detection has become a critical challenge. A wide variety of models is available, but inconsistent datasets, evaluation metrics, and assessment strategies obscure comparisons of model effectiveness. To address this, we evaluate 15 different detection models from six distinct systems, as well as seven trained models, across seven English-language textual test sets and three creative human-written datasets. We provide an empirical analysis of model performance, the influence of training and evaluation data, and the impact of key metrics. We find that no single system excels in all areas and nearly all are effective for certain tasks, and the representation of model performance is critically linked to dataset and metric choices. We find high variance in model ranks based on datasets and metrics, and overall poor performance on novel human-written texts in high-risk domains. Across datasets and metrics, we find that methodological choices that are often assumed or overlooked are essential for clearly and accurately reflecting model performance. 
 \\ \newline \Keywords{machine-generated text detection, deepfake detection, evaluation, metrics} }

\begin{document}

\maketitleabstract

\section{Introduction}
    Recent years have witnessed remarkable advancements in generative systems capable of producing realistic video, audio, and text. While these technologies offer significant benefits, they also introduce serious challenges in verifying the origin of content. Distinguishing between machine-generated and human-written text, in particular, has become increasingly critical. In response, research into machine-generated text detection\footnote{As a terminological note, we prefer the term "machine-generated" to "deepfake," as there may not be any intent to deceive. This also mirrors the language of the datasets used.} -- identifying and sourcing content produced by generative systems -- has grown rapidly.

    Machine-generated text detection is essential across domains such as education, media, and security, with use cases including the mitigation of spam, plagiarism, fraud, propaganda, and more \cite{rosca2025,lee2023,saravani2021}. Modern large language models (LLMs) pose unique challenges, as even humans have difficulty detecting machine-generated text \cite{lee-2025,uchende2023}. Evaluation practices are inconsistent, with significant variations in datasets, metrics, and methodologies, and the rationale behind these methodological choices is often unclear. While previous research has explored the effects of data and metrics \cite{pudasaini-2025,bhattacharjee-2024,zhang-2023}, the scope of this work is limited in scope based on models, datasets, metrics, and depth of analysis.

    To address this gap, we examine 15 detection variants across six systems, alongside pretrained transformer- and feature-based models. Our work uniquely takes a deeper look at the dataset and metric attributes that are often overlooked, and in doing so, brings to light essential issues in model performance and development. Our analysis uncovers findings in two main areas:

    \begin{itemize}
    \item \textbf{Data}: Model performance varies substantially depending on the evaluation data: across four datasets, F1 scores range from approximately 0 to 0.982, with nearly all models performing well on certain datasets and weak on others (Section \ref{sec:baseline}). With regard to training, fine-tuning on in-domain data yields models that outperform zero-shot and externally trained public models, though for some datasets, out-of-domain training yields better results (Section \ref{sec:cross}). On three novel human-written datasets, all but three variants exhibit an error rate of at least a 15\% on one or more datasets, while those with lower error rates also suffered from low recall across all datasets. This highlights potential risks in real-world deployment (Section \ref{sec:novel}).
        
    \item \textbf{Metrics}: Metrics can be exploited to misrepresent model performance: common metrics like F1 score, area under the receiver operating curve (AUROC), and true positive rate at false positive rate 1\% (TPR@FPR 1\%) pattern differently depending on the label distribution of the evaluation set, which are frequently imbalanced (Section \ref{sec:balance}). Over eight commonly used metrics, model rankings variances ranges from 0.77 to 15.25 over 15 models, with the choice of classification threshold having a significant impact (Section \ref{sec:ranking}).
    \end{itemize}

    We conclude with an analysis of potential causes of model disparity, including input length, punctuation, repetition, and perplexity. Our analysis shines a light on dataset and metric-related aspects of the task that are often overlooked. We establish the necessity of using multiple, well-motivated metrics and datasets for evaluation, as errors can be obfuscated and model performance misinterpreted depending on evaluation settings.
    
\section{Systems}
    \label{sec:systems}

\begin{table*}[h!]
\centering
\small
\begin{threeparttable}
\def\arraystretch{.95}
\setlength\tabcolsep{5.0pt}
\begin{tabular}{llll}
\textbf{Zero-shot Models} & \textbf{Variants} & \textbf{Evaluation Data} & \textbf{Key Performance}\\
\midrule
\makecell[l]{Binoculars \\ \cite{hans2024}} & Falcon &  \makecell[l]{\newcite{verma2024}, \\\newcite{lian2023}, \\ custom}  &  \makecell[l]{TPR-FPR: 0.76 - 0.98, \\ F1: 0.985 - 0.994} \\ \hline
\multirow{3}{*}{\makecell[l]{Fast-DetectGPT \\ (\textsc{fdg}) \\ \cite{bao2023}}} & gpt-neo & \multirow{3}{*}{Custom} & \multirow{3}{*}{AUROC: 0.9754 - 0.9984} \\
    & gpt-j \\
    & falcon-7b \\ \hline
\multirow{2}{*}{\makecell[l]{Zippy \\ \cite{thinkst2023}}} & LZMA & \multirow{2}{*}{Various \tnote{a}} & \multirow{2}{*}{AUROC: 0.76 - 0.82} \\
\vspace{.5em}
    & Ensemble \\
\textbf{Trained Models} & \textbf{Variants} & \textbf{Evaluation Data} & \textbf{Key Performance} \\
\midrule
\makecell[l]{RADAR \\ \cite{hu2023}} & Base & Custom & AUROC: 0.763 - 0.955 \\ \hline
\multirow{4}{*}{\makecell[l]{BiScope \\ \cite{guo2024-1}}} & Arxiv & \multirow{4}{*}{Custom} & \multirow{4}{*}{F1: 0.5456 - 1.0}\\
    & Yelp \\
    & Essay \\
    & Creative \\ \hline
\multirow{4}{*}{\makecell[l]{DeTeCtive \\ \cite{guo2024}}} & MAGE (Deepfake) & MAGE (Deepfake) & \multirow{4}{*}{F1: 0.8260 - 0.9974} \\
    & M4 & M4 &\\
    & TuringBench & TuringBench \\
\vspace{.5em}
    & OUTFOX & OUTFOX \\ 
\textbf{\makecell[l]{Pretrained \\Transformers\tnote{b}}} & \textbf{Model} & \textbf{Evaluation Data} & \textbf{Key Performance}\\
\midrule
DistilBERT & \texttt{distilbert-base-cased} & N/A & N/A\\
BERT & \texttt{bert-base-cased} & N/A & N/A \\
RoBERTa & \texttt{roberta-base} & N/A & N/A \\
Longformer & \texttt{longformer-base-4096} & N/A & N/A \\
\vspace{.5em}DeBERTa & \texttt{deberta-v3-base} & N/A & N/A \\
\textbf{Feature-Based} & \textbf{Features} & \textbf{Evaluation Data} & \textbf{Key Performance} \\
\hline
\textsc{mcgovern} & \makecell[l]{Word, part-of-speech, \\ character n-grams} &  \makecell[l]{MAGE (Deepfake) \\ M4 \\ \newcite{guo2023} \\ OUTFOX \\ Ghostbuster} &  \makecell[l]{AUROC: 0.943 - 0.996 \\ F1: 0.947 - 0.987} \\ \hline
\textsc{stylo}\tnote{c} & Linguistic/stylometric & N/A & N/A \\\\
\end{tabular}
\begin{tablenotes}
    \item[a] See \url{https://blog.thinkst.com/2023/06/meet-zippy-a-fast-ai-llm-text-detector.html}
    \item[b] While previous work has included transformer fine-tuning, there is no consensus on dataset, metrics, or performance.
    \item[c] This system is implemented independent of any specific previous work.
\end{tablenotes}
\end{threeparttable}
\caption{Summary of the models and variants evaluated. We report the main datasets and metrics used, along with the range of values for the given system for the key metric(s).}
\label{tab:models}
\end{table*}


We assemble a diverse collection of contemporary machine-generated text detection systems, including zero-shot and trained public models, pretrained transformers, and feature-based approaches. Table \ref{tab:models} summarizes the selected model variants, detailing their original evaluation data, metrics, and reported performance. Our goal is to explore impacts of datasets and metrics through rigorous evaluation and analysis as in \newcite{wu2025}; we build on their work by including more models, as well as a deeper analysis of performance as it relates to metrics and datasets.

    \subsection{Public Models}
We explore both zero-shot and trained models that are publicly available. We focus on detection models that satisfy three key criteria: (1) contemporary (since 2023), (2) fully open-source (enabling transparency and customization), and (3) free to use (excluding systems dependent on proprietary APIs like OpenAI, which incur additional expenses and requirements). We prioritize transparent, easily implementable systems with minimal operational costs. These systems all report strong performance, with reported F1/AUROC scores over 0.95 on many of their respective datasets. 

    \subsection{Pretrained Transformers}
We include in our evaluation fine-tuning of five transformer-based architectures of varying scales and configurations: DistilBERT \cite{shah2023}, BERT \cite{devlin2019}, RoBERTa \cite{zhuang2021}, Longformer \cite{beltagy2020}, and DeBERTa \cite{he2021}. Notably, prior work has shown RoBERTa and Longformer to be particularly effective for machine-generated detection \cite{apollo2025,li2024,pu2023}. We fine-tune and evaluate each of these models on a shared dataset, enabling direct comparison with the above approaches.
    
    \subsection{Feature-based Models}
    Feature-based models have demonstrated notable effectiveness in machine-generated text detection, leveraging the distinct linguistic and stylometric patterns characteristic of different LLMs \cite{mcgovern2025,munoz2024}. We implement two variations: (1) a feature-based classifier based on the work of \newcite{mcgovern2025} (\textsc{mcgovern}), which uses a combination word, part-of-speech, and character n-grams, trained with a gradient boosting classifier, and (2) a series of linguistic and stylometric features adapted from \newcite{almazrouei2023} (\textsc{stylo}). This model combines linguistic and stylometric features to train an ensemble of classifiers. 
    \footnote{Implementation details for all systems in Appendix \ref{app:systems}.}

    \subsection{Other models}
    We aim for a diverse set of models based on the above criteria, but many other models are available. Some were excluded for targeting specific generative models (e.g., GPT) or datasets \cite{venkatraman2024,koike2024,wu2023,yang2023,gehrmann2019}. Others were built on free models from OpenAI which are no longer available or require restricted model access \cite{bao2025,verma2024,mao2024}: we restrict our evaluation to only free and open models. We also exclude the many commercial systems for similar reasons, as well as the inherent opacity of proprietary models.

\section{Data}
\label{sec:data}
To ensure standardized, robust evaluations, we establish a unified dataset comprising seven test sets derived from four benchmark datasets:

\begin{figure*}[htbp]
\centerline{\includegraphics[width=\textwidth]{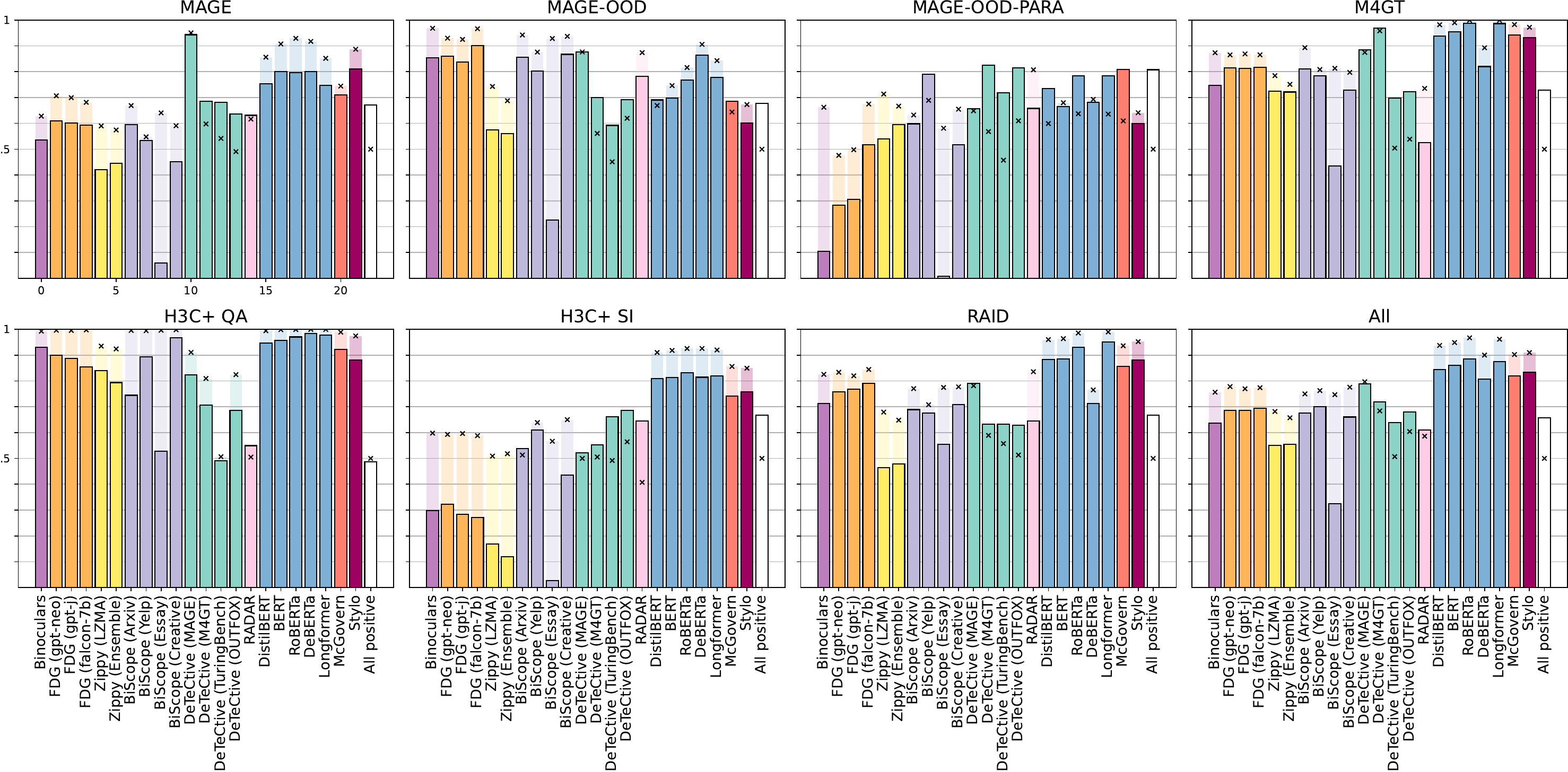}}
\caption{F1 and AUC (marked with \texttt{x}) for each model on each dataset. For the pretrained transformers, these were trained on in-domain training data from the respective dataset.}
\label{fig:baseline} 
\end{figure*}

    \paragraph{MAGE (aka Deepfake) \cite{li2024}:} This dataset consists of a 447k human and AI-generated text samples covering diverse models and methodologies. We extract three test sets: (1) a class-balanced sample of 10k from the full dataset (\mage{}); and (2-3) the two "wilder" test sets, one for out-of-domain texts (\mageood) and one for out-of-domain paraphrases (\mageoodpara).

    \paragraph{RAID \cite{dugan2024}:} 
    The RAID corpus comprises 11 million human-written and machine-generated texts with adversarial examples crafted to evade detection systems. We extract a class-balanced subset of 10k samples from the RAID training partition (\raid), as the official test set labels are withheld for leaderboard purposes.
    
    \paragraph{H3C+ Corpus \cite{su2024}:} This corpus extends the H3C corpus \cite{guo2023} with semantic-invariant perturbations, shown to be more difficult to detect. We utilize their two English test sets, sampling 10k class-balanced texts each from the English partitions for question answering (\hcqa) and semantic invariance (\hcsi).

    \paragraph{M4GT-Bench \cite{wang2024}:} A multilingual, multidomain corpus covering several generated methods. We take a class-balanced sample of 10k texts from the English partition of their test corpus (\mgt).

    \vspace{.3em}
    
These datasets were chosen due to (1) public availability, (2) diversity in models and domains, and (3) common usage. They each contain many models used for generation, different domains, and different styles of text. There are many other high-quality datasets available \cite{wu2025,koike2024,verma2024,liu2024,lian2023,uchendu2021}; see \newcite{gritsai-2025} for a contemporary overview of datasets and resulting challenges in detection. Our goal is not to comprehensively evaluate models against all datasets, but rather highlight disparities in performance across representative datasets.

We conduct comprehensive evaluations across all seven test sets, including an aggregated analysis of the combined datasets (\all). Some public models (e.g., DeTeCtive (MAGE)) were trained on other partitions from these resources: before evaluation, we have ensured that there is no overlap between the training data and our evaluation sets. For our training, we create training sets of 10k texts for each dataset by random sampling from the respective training partitions, ensuring no overlap with the evaluation data.

\section{Experiments}
\label{sec:baseline}

We evaluate each system on each of the eight datasets described above. For comparison include a trivial baseline system (All positive) that assigns every sample a score of 1 (machine-generated).\footnote{Notation varies among the datasets about which class is machine-generated; we ensure that all datasets and models are normalized to have 1 as the machine-generated class.} For our initial evaluation, we use F1 score (threshold = 0.5) and AUROC; we discuss more on metrics in Section \ref{sec:metrics}. Results are shown in Figure \ref{fig:baseline}.

We start by assessing some key trends in our evaluation. First, performance varies substantially across models, with no system dominating uniformly. Even architecturally similar models exhibit irregular performance patterns.

Among public models, DeTeCtive (MAGE) achieves the highest aggregate F1/AUROC scores, though Binoculars, BiScope, Fast-DetectGPT, RADAR, and Zippy perform competitively on specific test sets. In several cases, the "All positive" baseline outperforms many systems, particularly with regard to F1 score (e.g., on the \mageood{} dataset). As some datasets are heavily weighted towards machine-generated content, this baseline is deceptively strong, and underscores the limitations of F1 score as a metric (Section \ref{sec:metrics}). 

\begin{figure*}[t!]
\includegraphics[width=\textwidth]{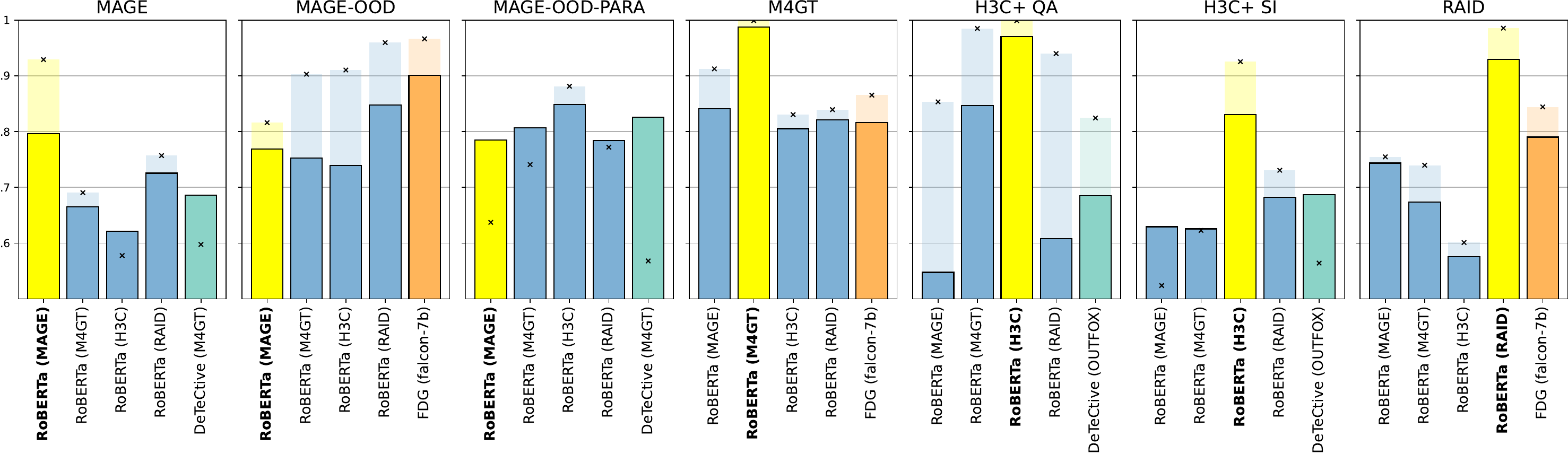}
\caption{F1 and AUROC scores for cross-trained models. Bold models are those trained on the in-domain dataset. Each plot reflects a dataset, with the best-performing public model on that dataset included for reference.}
\label{fig:cross} 
\end{figure*}

    \begin{table*}[h]
\centering
\small
\begin{tabular}{l|cc|cc||c}
\textbf{Model} & \textbf{Low} & \textbf{Dataset} & \textbf{High} & \textbf{Dataset} & \textbf{Mean}\\
\hline
Binoculars & \cellcolor{green!10} 0.60 & \mageoodpara{} & \cellcolor{green!49} 0.99 & \hcqa{} & \cellcolor{green!29} 0.79\\
FDG (gpt-neo) & \cellcolor{green!00} 0.48 & \mageoodpara{} & \cellcolor{green!50} 1.00 & \hcqa{} & \cellcolor{green!27} 0.77\\
FDG (gpt-j) & \cellcolor{green!00} 0.50 & \hcsi{} & \cellcolor{green!50} 1.00 & \hcqa{} & \cellcolor{green!27} 0.77\\
FDG (falcon-7b) & \cellcolor{green!09} 0.59 & \hcsi{} & \cellcolor{green!50} 1.00 & \mageood{} & \cellcolor{green!30} 0.80\\
Zippy (LZMA) & \cellcolor{green!01} 0.51 & \hcsi{} & \cellcolor{green!43} 0.93 & \hcqa{} & \cellcolor{green!21} 0.71\\
Zippy (Ensemble) & \cellcolor{green!02} 0.52 & \hcsi{} & \cellcolor{green!42} 0.92 & \hcqa{} & \cellcolor{green!18} 0.68\\
BiScope (Arxiv) & \cellcolor{green!01} 0.51 & \hcsi{} & \cellcolor{green!50} 1.00 & \mageood{} & \cellcolor{green!27} 0.77\\
BiScope (Yelp) & \cellcolor{green!05} 0.55 & \mage{} & \cellcolor{green!49} 0.99 & \hcqa{} & \cellcolor{green!25} 0.75\\
BiScope (Essay) & \cellcolor{green!07} 0.57 & \mageoodpara{} & \cellcolor{green!50} 1.00 & \raid{} & \cellcolor{green!26} 0.76\\
BiScope (Creative) & \cellcolor{green!09} 0.59 & \hcsi{} & \cellcolor{green!50} 1.00 & \hcqa{} & \cellcolor{green!27} 0.77\\
DeTeCtive (MAGE) & \cellcolor{green!00} 0.50 & \hcsi{} & \cellcolor{green!45} 0.95 & \mage{} & \cellcolor{green!29} 0.79\\
DeTeCtive (M4GT) & \cellcolor{green!01} 0.51 & \hcsi{} & \cellcolor{green!46} 0.96 & \mgt{} & \cellcolor{green!16} 0.66\\
DeTeCtive (TuringBench) & \cellcolor{green!00} 0.45 & \hcqa{} & \cellcolor{green!06} 0.56 & \mageoodpara{} & \cellcolor{green!00} 0.50\\
DeTeCtive (OUTFOX) & \cellcolor{green!00} 0.49 & \raid{} & \cellcolor{green!32} 0.82 & \mageoodpara{} & \cellcolor{green!09} 0.59\\
RADAR & \cellcolor{green!00} 0.41 & \mgt{} & \cellcolor{green!37} 0.87 & \mageood{} & \cellcolor{green!18} 0.68\\
\end{tabular}
\caption{Lowest and highest reported AUROC scores for the best performing variant of each model across the seven unique datasets.}
\label{tab:differences}
\end{table*}

Models exhibit significant variability in performance on their best and worst datasets as well, as detailed in Table \ref{tab:differences}. Binoculars has the highest minimum AUROC (0.60): \textit{every} model explored performs near chance ($\leq$0.60) on at least one of the evaluation sets, yet these models all perform strongly on others. Excepting the TuringBench DeTeCtive variant, all models scored $\geq$0.82 AUROC on their best dataset. This underscores the importance of diverse evaluation data: models can easily be over- or under-represented best on their performance on different datasets.

Pretrained models outperform publicly available systems in this setting, in which they are trained on in-domain data. The transformer variants all exhibit similar performance. DeBERTa exhibits higher variance, while the other four models are consistent. Feature-based models, both the base n-grams of \textsc{mcgovern} and the more complicated stylometric features, are notably strong, achieving performance comparable to transformers in many cases. We assess the ability of models trained in a cross-domain setting in Section \ref{sec:cross}.

Performance disparities are stark: while most models excel on the \hcqa{} dataset, the \hcsi{} and \mageoodpara{} datasets prove challenging across all models. This is critical when comparing and evaluating models: the choice of evaluation dataset is highly predictive of model performance.

\subsection{Cross-Training}
\label{sec:cross}

In order to evaluate the robustness of these trained models to new datasets, we conduct cross-training analysis, where models are trained on one dataset and evaluated on others. We train four models using 10k samples from each of the above datasets (MAGE, M4GT, H3C+, and RAID), and evaluate on the previous defined test sets. MAGE-trained models treat the three MAGE-based sets as in-domain; H3C+-trained models treat the two H3C+-based sets as in-domain; similarly for RAID and M4GT; the evaluation datasets which don't match the training data are considered out-of-domain.  Results are shown in Figure \ref{fig:cross}.
    
As expected, in-domain performance is substantially better than out-of-domain for most cases, with a mean AUROC score over all datasets 0.06 above the best out-of-domain model and better on five of the seven datasets. Models capture dataset-specific generation patterns but struggle to generalize to adversarial or out-of-domain examples. This is especially evident in the RAID dataset, which contains more specific adversarial attacks.

However, we find in some cases out-of-domain performance can exceed in-domain: for the \mageood{} and \mageoodpara{} sets, models trained on other datasets perform much better than even those trained on MAGE data, indicating that potential challenges can be mitigated by incorporating other external datasets into training.  Out-of-domain models even outperform the best public models on both \mage{} and \mgt{}.

\subsection{Novel Human-Written Texts}
\label{sec:novel}
False positives in machine-generated text detection carry substantial risk for harm, include educational impacts and unfair censorship \cite{wu2025}. To further explore model performance, we evaluate models on datasets comprising exclusively human-authored texts:

\vspace{-.3em}
    \paragraph{Stories in the Wild:} \cite{august2020}: This dataset contains 1,630 samples of human-written creative fiction.
\vspace{-.3em}
    \paragraph{PERSUADE:} \cite{crossley2024}: This corpus comprises 25k persuasive essays by secondary students (ages 11-17); we evaluate on a 5k random subset.
\vspace{-.3em}
    \paragraph{Huang:} \cite{huang2020}: Derived from this multi-lingual dataset built for hate speech analysis on X (formerly Twitter), we aggregate 5k  user profiles from the English partition of the dataset by combining all of the users' posts into a single text. This dataset is particularly challenging, as many of the texts are quite short (mean length of 26 words) and contain informal language and punctuation.

\vspace{.3em}
We selected these datasets to reflect high-stakes scenarios. \newcite{august2020} and \newcite{huang2020} focus on creative works in different domains, both of which could potentially be subject to censorship and/or unfair rejection if detected as machine-generated. \newcite{crossley2024} reflects student work, where the use of generative tools for authorship is often disallowed, and students are susceptible to negative consequences and punitive measures if their work is incorrectly identified as being generated (or assisted) by automated tools. Additionally, these datasets are useful in their disparate domains and styles: each contains very different styles, formatting, length, and other attributes which are potentially difficult for machine-generated text detction systems.

We report accuracy (equal to the true negative rate) on each of these three datasets in Table \ref{tab:human}, as well as the mean. These are contrasted with recall scores on the data from Section \ref{sec:baseline}: ideal models should maximize both metrics, with high accuracy of human-written texts (minimal false positives) while maintaining high recall on machine-generated data (minimal false negatives).\footnote{We use RoBERTa as an exemplar for fine-tuned models; the others had similar performance.}

\begin{table}[t!]
\small
\setlength\tabcolsep{2.5pt}
\begin{tabular}{l|ccc|c|c}
   \textbf{System} & \rotatebox{90}{\textbf{Stories}} & \rotatebox{90}{\textbf{PERSUADE}} & \rotatebox{90}{\textbf{Huang}} & \rotatebox{90}{\textbf{Mean Acc.}} & \rotatebox{90}{\textbf{Recall (\all)}} \\
   \midrule
Binoculars & 0.90 & 1.00 & 0.97 & \cellcolor{green!46} 0.96 & \cellcolor{green!00} 0.50 \\
BiScope-Arxiv & 0.86 & 0.99 & 0.82 & \cellcolor{green!39} 0.89 & \cellcolor{green!03} 0.53 \\
BiScope-Yelp & 0.85 & 1.00 & 0.98 & \cellcolor{green!44} 0.94 & \cellcolor{green!11} 0.61 \\
BiScope-Creative & 1.00 & 1.00 & 1.00 & \cellcolor{green!50} 1.00 & \cellcolor{green!00} 0.41 \\
BiScope-Essay & 1.00 & 1.00 & 1.00 & \cellcolor{green!50} 1.00 & \cellcolor{green!00} 0.26 \\
DeTeCtive (MAGE) & 0.58 & 0.88 & 0.32 & \cellcolor{green!09} 0.59 & \cellcolor{green!33} 0.83 \\
DeTeCtive (M4GT) & 0.45 & 0.93 & 0.63 & \cellcolor{green!17} 0.67 & \cellcolor{green!35} 0.85 \\
DeTeCtive (OUTFOX) & 0.70 & 0.92 & 0.33 & \cellcolor{green!15} 0.65 & \cellcolor{green!43} 0.93 \\
DeTeCtive (TuringBench) & 0.06 & 0.03 & 0.00 & \cellcolor{green!00} 0.03 & \cellcolor{green!40} 0.90 \\
\textsc{fdg} (gpt-neo) & 0.85 & 0.85 & 0.83 & \cellcolor{green!34} 0.84 & \cellcolor{green!11} 0.61 \\
\textsc{fdg} (gpt-j) & 0.86 & 0.79 & 0.79 & \cellcolor{green!32} 0.82 & \cellcolor{green!13} 0.63 \\
\textsc{fdg} (falcon-7b) & 0.75 & 0.83 & 0.96 & \cellcolor{green!35} 0.85 & \cellcolor{green!13} 0.63 \\
RADAR & 0.18 & 0.68 & 0.00 & \cellcolor{green!00} 0.29 & \cellcolor{green!16} 0.66 \\
Zippy (LZMA) & 0.95 & 0.34 & 0.96 & \cellcolor{green!25} 0.75 & \cellcolor{green!00} 0.43 \\
Zippy (Ensemble) & 0.91 & 0.20 & 0.96 & \cellcolor{green!19} 0.69 & \cellcolor{green!00} 0.46 \\    \midrule
RoBERTa (H3C+) & 0.53 & 0.69 & 1.00 & \cellcolor{green!24} 0.74 & \cellcolor{green!33} 0.83 \\
RoBERTa (M4GT) & 0.74 & 0.93 & 0.50 & \cellcolor{green!22} 0.72 & \cellcolor{green!32} 0.82 \\
RoBERTa (MAGE) & 0.37 & 0.54 & 0.03 & \cellcolor{green!00} 0.31 & \cellcolor{green!41} 0.91 \\
RoBERTa (RAID) & 0.32 & 0.04 & 0.15 & \cellcolor{green!00} 0.17 & \cellcolor{green!41} 0.91 \\
\textsc{stylo} (H3C+) & 0.84 & 0.86 & 1.00 & \cellcolor{green!40} 0.90 & \cellcolor{green!02} 0.52 \\
\textsc{stylo} (M4GT) & 0.12 & 0.48 & 0.19 & \cellcolor{green!00} 0.26 & \cellcolor{green!38} 0.88 \\
\textsc{stylo} (MAGE) & 0.48 & 0.35 & 0.56 & \cellcolor{green!00} 0.46 & \cellcolor{green!27} 0.77 \\
\textsc{stylo} (RAID) & 0.35 & 0.70 & 0.02 & \cellcolor{green!00} 0.36 & \cellcolor{green!30} 0.80 \\
\textsc{mcgovern} (H3C+) & 0.21 & 0.15 & 1.00 & \cellcolor{green!00} 0.45 & \cellcolor{green!35} 0.85 \\
\textsc{mcgovern} (M4GT) & 0.11 & 0.23 & 0.00 & \cellcolor{green!00} 0.11 & \cellcolor{green!46} 0.96 \\
\textsc{mcgovern} (MAGE) & 0.04 & 0.01 & 0.00 & \cellcolor{green!00} 0.02 & \cellcolor{green!48} 0.98 \\
\textsc{mcgovern} (RAID) & 0.35 & 0.38 & 0.00 & \cellcolor{green!00} 0.25 & \cellcolor{green!36} 0.86 \\
\bottomrule
    \end{tabular}
\caption{Accuracy on novel human datasets, along the recall scores from the original evaluation  (Section \ref{sec:baseline})}
\label{tab:human}
\end{table}

On these datasets we again observe high performance variability across models. Models that perform well on this task tend to have worse recall on the original data: they simply make fewer positive predictions. No system simultaneously exceeds 0.8 accuracy and 0.8 recall. 

Binoculars, Zippy, BiScope and \textsc{fdg} variants maintain strong accuracy, but suffer from poor recall. DeTeCtive variants and RADAR have higher recall but low accuracy. All other models fall below 85\% accuracy on at least one dataset. Some models perform well only on certain datasets: \textsc{fdg} (falcon-7b) is highly accurate on the Huang dataset but worse on the others, while Zippy is strong on both Stories and PERSUADE datasets but classifies more than half of the Huang dataset as machine-generated. Trained models prioritize recall the expense of accuracy, and thus perform poorly, although we again see impacts of training data: models trained on the H3C+ corpus tend to have much better balance.

Our experiments reveal a significant vulnerability: models struggle with unseen creative human-written texts. Standard evaluations using imbalanced datasets (heavily skewed toward machine-generated content) likely inflate perceived performance, masking this weakness. Most systems either pose substantial risk when applied to out-of-domain human writing or struggle with recall, requiring careful deployment consideration.

Models have highly variable performance based on the evaluation data used, which could highlight or obscure aspects of performance. We now turn to metrics, which can have similar effects.

    \begin{table*}[h]
\centering
\small
\setlength\tabcolsep{.1pt}
\begin{tabular}{l|ccccc|ccccc|ccc|c}
\toprule
& \multicolumn{5}{c|}{Threshold 0.5} & \multicolumn{5}{c|}{Threshold by EER} & \multicolumn{3}{c|}{Threshold invariant} & \\
Model & \rotatebox{90}{Precision} & \rotatebox{90}{Recall} & \rotatebox{90}{F1} & \rotatebox{90}{Accuracy} & \rotatebox{90}{AvgRec} & \rotatebox{90}{Precision} & \rotatebox{90}{Recall} & \rotatebox{90}{F1} & \rotatebox{90}{Accuracy} & \rotatebox{90}{AvgRec} & \rotatebox{90}{AUROC} & \rotatebox{90}{TPR@FPR 1\%} & \rotatebox{90}{TPR@FPR .01\%} & \rotatebox{90}{Variance}\\ 
\midrule
Binoculars & \cellcolor{green!30} 2.57 & \cellcolor{green!00} 11.00 & \cellcolor{green!09} 7.86 & \cellcolor{green!17} 5.71 & \cellcolor{green!21} 4.86 & \cellcolor{green!00} 11.71 & \cellcolor{green!36} 1.00 & \cellcolor{green!11} 7.14 & \cellcolor{green!00} 10.57 & \cellcolor{green!00} 11.71 & \cellcolor{green!24} 4.00 & \cellcolor{green!19} 5.29 & \cellcolor{green!22} 4.57 & \cellcolor{red!47} 11.77\\
FDG (gpt-neo) & \cellcolor{green!14} 6.43 & \cellcolor{green!05} 8.71 & \cellcolor{green!15} 6.14 & \cellcolor{green!20} 5.00 & \cellcolor{green!19} 5.29 & \cellcolor{green!18} 5.57 & \cellcolor{green!00} 11.29 & \cellcolor{green!12} 7.00 & \cellcolor{green!21} 4.71 & \cellcolor{green!21} 4.71 & \cellcolor{green!18} 5.43 & \cellcolor{green!28} 3.00 & \cellcolor{green!18} 5.57 & \cellcolor{red!16} 3.92\\
FDG (gpt-j) & \cellcolor{green!09} 7.86 & \cellcolor{green!07} 8.29 & \cellcolor{green!11} 7.14 & \cellcolor{green!15} 6.29 & \cellcolor{green!14} 6.57 & \cellcolor{green!13} 6.71 & \cellcolor{green!00} 10.71 & \cellcolor{green!09} 7.71 & \cellcolor{green!16} 6.00 & \cellcolor{green!16} 6.00 & \cellcolor{green!19} 5.29 & \cellcolor{green!27} 3.29 & \cellcolor{green!15} 6.14 & \cellcolor{red!11} 2.79\\
FDG (falcon-7b) & \cellcolor{green!19} 5.14 & \cellcolor{green!09} 7.86 & \cellcolor{green!15} 6.29 & \cellcolor{green!21} 4.71 & \cellcolor{green!25} 3.86 & \cellcolor{green!25} 3.86 & \cellcolor{green!00} 10.43 & \cellcolor{green!12} 7.00 & \cellcolor{green!22} 4.57 & \cellcolor{green!26} 3.57 & \cellcolor{green!31} 2.14 & \cellcolor{green!29} 2.71 & \cellcolor{green!26} 3.57 & \cellcolor{red!20} 4.89\\
Zippy (LZMA) & \cellcolor{green!13} 6.71 & \cellcolor{green!00} 13.00 & \cellcolor{green!00} 11.57 & \cellcolor{green!05} 8.86 & \cellcolor{green!06} 8.43 & \cellcolor{green!00} 12.71 & \cellcolor{green!32} 2.00 & \cellcolor{green!07} 8.14 & \cellcolor{green!00} 11.57 & \cellcolor{green!00} 12.71 & \cellcolor{green!09} 7.71 & \cellcolor{green!07} 8.29 & \cellcolor{green!02} 9.43 & \cellcolor{red!35} 8.70\\
Zippy (Ensemble) & \cellcolor{green!03} 9.14 & \cellcolor{green!00} 11.71 & \cellcolor{green!00} 12.00 & \cellcolor{green!00} 10.00 & \cellcolor{green!01} 9.86 & \cellcolor{green!00} 13.71 & \cellcolor{green!28} 3.00 & \cellcolor{green!03} 9.14 & \cellcolor{green!00} 12.57 & \cellcolor{green!00} 13.71 & \cellcolor{green!04} 9.00 & \cellcolor{green!02} 9.43 & \cellcolor{green!00} 10.29 & \cellcolor{red!28} 7.08\\
BiScope (Arxiv) & \cellcolor{green!05} 8.71 & \cellcolor{green!13} 6.71 & \cellcolor{green!12} 7.00 & \cellcolor{green!09} 7.71 & \cellcolor{green!11} 7.29 & \cellcolor{green!12} 7.00 & \cellcolor{green!02} 9.43 & \cellcolor{green!09} 7.71 & \cellcolor{green!11} 7.29 & \cellcolor{green!15} 6.29 & \cellcolor{green!13} 6.86 & \cellcolor{green!11} 7.14 & \cellcolor{green!15} 6.14 & \cellcolor{red!03} 0.77\\
BiScope (Yelp) & \cellcolor{green!00} 10.43 & \cellcolor{green!22} 4.43 & \cellcolor{green!14} 6.43 & \cellcolor{green!13} 6.86 & \cellcolor{green!11} 7.29 & \cellcolor{green!09} 7.86 & \cellcolor{green!10} 7.43 & \cellcolor{green!11} 7.14 & \cellcolor{green!17} 5.86 & \cellcolor{green!15} 6.14 & \cellcolor{green!06} 8.43 & \cellcolor{green!02} 9.43 & \cellcolor{green!10} 7.57 & \cellcolor{red!09} 2.20\\
BiScope (Essay) & \cellcolor{green!24} 4.00 & \cellcolor{green!00} 14.71 & \cellcolor{green!00} 14.57 & \cellcolor{green!00} 12.14 & \cellcolor{green!00} 11.57 & \cellcolor{green!26} 3.57 & \cellcolor{green!00} 15.00 & \cellcolor{green!00} 14.43 & \cellcolor{green!00} 10.29 & \cellcolor{green!03} 9.29 & \cellcolor{green!10} 7.43 & \cellcolor{green!13} 6.71 & \cellcolor{green!11} 7.14 & \cellcolor{red!61} 15.25\\
BiScope (Creative) & \cellcolor{green!17} 5.86 & \cellcolor{green!04} 9.00 & \cellcolor{green!12} 7.00 & \cellcolor{green!15} 6.29 & \cellcolor{green!17} 5.71 & \cellcolor{green!22} 4.43 & \cellcolor{green!00} 11.43 & \cellcolor{green!10} 7.57 & \cellcolor{green!18} 5.43 & \cellcolor{green!22} 4.43 & \cellcolor{green!21} 4.71 & \cellcolor{green!14} 6.43 & \cellcolor{green!17} 5.71 & \cellcolor{red!14} 3.58\\
DeTeCtive (MAGE) & \cellcolor{green!11} 7.14 & \cellcolor{green!23} 4.29 & \cellcolor{green!25} 3.86 & \cellcolor{green!21} 4.71 & \cellcolor{green!23} 4.29 & \cellcolor{green!18} 5.43 & \cellcolor{green!11} 7.14 & \cellcolor{green!22} 4.43 & \cellcolor{green!21} 4.71 & \cellcolor{green!26} 3.57 & \cellcolor{green!00} 10.00 & \cellcolor{green!00} 10.43 & \cellcolor{green!11} 7.29 & \cellcolor{red!19} 4.79\\
DeTeCtive (M4GT) & \cellcolor{green!00} 10.43 & \cellcolor{green!24} 4.00 & \cellcolor{green!17} 5.86 & \cellcolor{green!07} 8.29 & \cellcolor{green!02} 9.43 & \cellcolor{green!09} 7.71 & \cellcolor{green!12} 7.00 & \cellcolor{green!16} 6.00 & \cellcolor{green!14} 6.43 & \cellcolor{green!11} 7.29 & \cellcolor{green!00} 12.00 & \cellcolor{green!00} 11.71 & \cellcolor{green!00} 10.57 & \cellcolor{red!22} 5.58\\
DeTeCtive (TuringBench) & \cellcolor{green!00} 13.86 & \cellcolor{green!25} 3.71 & \cellcolor{green!06} 8.43 & \cellcolor{green!00} 12.86 & \cellcolor{green!00} 14.29 & \cellcolor{green!00} 12.14 & \cellcolor{green!13} 6.71 & \cellcolor{green!02} 9.43 & \cellcolor{green!00} 12.00 & \cellcolor{green!00} 12.57 & \cellcolor{green!00} 14.00 & \cellcolor{green!00} 13.57 & \cellcolor{green!00} 14.43 & \cellcolor{red!41} 10.24\\
DeTeCtive (OUTFOX) & \cellcolor{green!00} 13.14 & \cellcolor{green!27} 3.29 & \cellcolor{green!10} 7.57 & \cellcolor{green!00} 10.86 & \cellcolor{green!00} 11.86 & \cellcolor{green!00} 10.57 & \cellcolor{green!15} 6.14 & \cellcolor{green!09} 7.86 & \cellcolor{green!03} 9.14 & \cellcolor{green!01} 9.86 & \cellcolor{green!00} 12.71 & \cellcolor{green!00} 12.29 & \cellcolor{green!00} 12.43 & \cellcolor{red!32} 7.98\\
RADAR & \cellcolor{green!06} 8.57 & \cellcolor{green!03} 9.29 & \cellcolor{green!07} 8.29 & \cellcolor{green!01} 9.71 & \cellcolor{green!02} 9.43 & \cellcolor{green!12} 7.00 & \cellcolor{green!00} 11.29 & \cellcolor{green!03} 9.29 & \cellcolor{green!05} 8.86 & \cellcolor{green!07} 8.14 & \cellcolor{green!00} 10.29 & \cellcolor{green!00} 10.29 & \cellcolor{green!03} 9.14 & \cellcolor{red!04} 1.10\\
\bottomrule
\end{tabular}
\caption{Mean rank (1-15) of each public model over the seven datasets for each metric.}
\label{tab:ranks}
\end{table*}

\section{Metrics}
    \label{sec:metrics}

    Our dataset analysis reveals divergences between F1 and AUROC metrics: while Binoculars, BiScope, and Fast-DetectGPT achieve strong AUROC scores despite comparatively low F1 scores, DeTeCtive models show an inverse relationship. We examine this relationship between metrics further with regard to class imbalance and model ranks.

    \subsection{Class Imbalance}
    \label{sec:balance}
    To assess metric sensitivity to class imbalance, we adjust the \mage{} dataset to contain arbitrary percentages of machine-generated data. We find that F1 scores increase logarithmically as the percentage of machine-generated content increases. F1 scores can thus be artificially inflated by predicting a high number of positive samples for a dataset that is skewed towards the positive class. Threshold independent metrics (AUROC, TPR) tend to remain stable regardless of the distribution of classes (see Appendix \ref{app:balance}).

    This partially explains the differences in model performance: models vary in relative strength when evaluated by different metrics. For a further comparative analysis, we study the impact of metrics on model ranks.

    \subsection{Ranking based on metrics}
    \label{sec:ranking}
    Model ranks likewise depend on the chosen metrics. We calculate the rank of each system (1 being the best, 15 the worst) across all seven test sets using a series of metrics. We evaluate precision, recall, F1 score, accuracy, and average recall. These metrics are binary and are calculate using a classification threshold. 
    
    We experiment with two thresholds: first, we use 0.5 which is considered the default. For some models, however, this threshold disproportionately favors precision or recall, as the models give very high or very low scores to most samples. As a comparison, we identified a model-specific threshold designed to optimize performance ("Threshold by EER").
    
    This threshold is calculated by first separating a sample of 1000 instances from each of the test corpora. We run each model on these samples, then identify the threshold that optimizes equal error rate (EER). This threshold is then used to classify the remainder of the test data; we report results on this subsample of the test data. 
    
    We also report results on threshold-independent metrics. We use model scores to calculate AUROC, TPR@FPR 1\%, and TPR@FPR 0.01\% 
    (See Appendix \ref{sec:metrics} for definitions of each metric). 
    Table \ref{tab:ranks} shows each model's mean rank (across models) and variance (across metrics).

    \begin{figure*}[t!]
    \includegraphics[width=\textwidth]{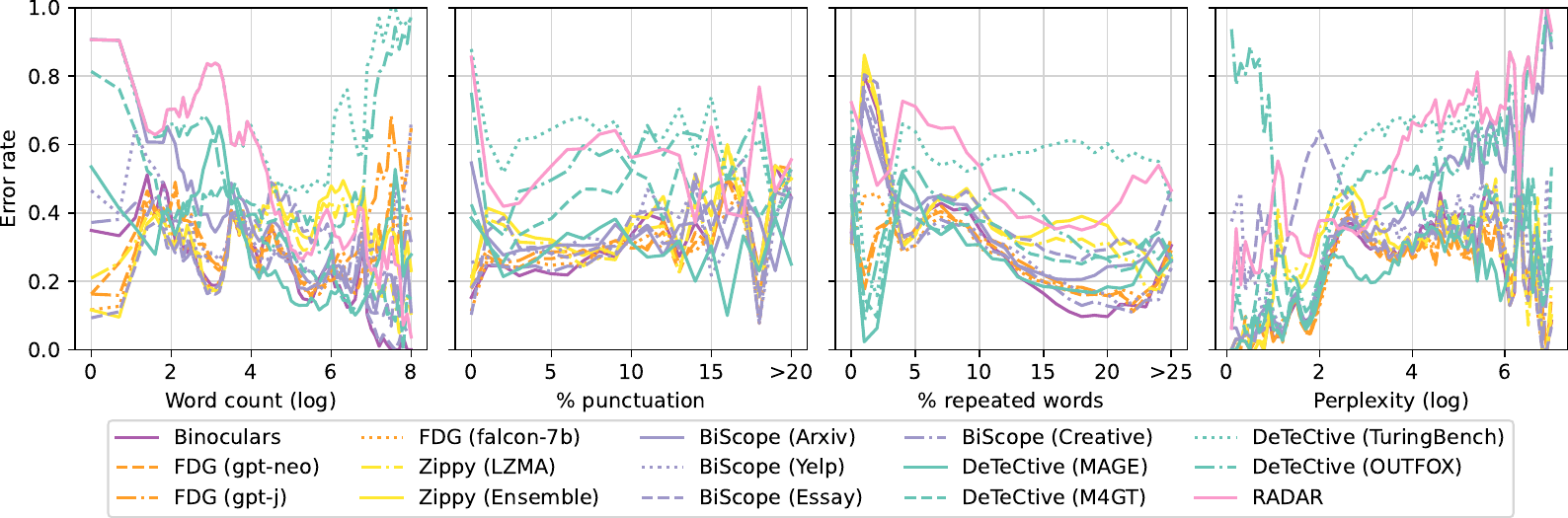}
    \caption{Model error as a function of word count (log), punctuation (\%), repeated words (\%), and perplexity (log).}
    \label{fig:analysis} 
    \end{figure*}

    The choice of metric can critically influences model ranks among competitors. At the 0.5 threshold, Binoculars and some BiScope models achieve high precision but low recall, DeTeCtive models exhibit high recall but underperform in AUROC and TPR@FPR scores, and \textsc{fdt} models have comparatively low F1 scores compared to other metrics. RADAR and BiScope (Arxiv) are stable but comparatively weak across metrics. Threshold adjustment dramatically alters model ranks: at an optimized threshold, Binoculars swaps from being high-precision to high-recall while FDG and Zippy models score much worse in recall. This highlights the necessity of diverse metrics: our understanding of system performance can vary drastically based on these small changes in metric calculations.

     These findings suggest two critical factors should be incorporated into metric choice: first, justified evaluation metric choices are essential. Most systems emphasize peak performance without explicitly justifying their metric choices,\footnote{With the notable exceptions being \newcite{bao2023}, who explicitly advocate for AUROC’s threshold-agnostic benefits, and \newcite{hans2024}, who prioritize TPR@FPR to minimize false positives in high-stakes scenarios.} but each metric captures distinct aspects that must be interpreted in context rather than in isolation. For instance, in student essay evaluation, where machine-generated content may be rare but false positives carry severe consequences, metrics emphasizing low false positive rates (e.g., TPR@FPR) are essential. In spam detection where synthetic content is prevalent and false positives are less risky, metrics like F1 may suffice. We find AUROC to be valuable in most scenarios, as it allows for better understanding of performance across multiple thresholds, which gives the end user better control.
     
    Second, a diverse set of metrics need to be used to balance potential performance differences stemming from the above findings. A more principled approach to metric selection, tailored to deployment contexts, is essential to ensure appropriate deployment of machine-generated  text detection.

\section{Analysis}

        To investigate performance differences, we explore four key textual attributes commonly used for machine-generated text detection: length (in words), punctuation, repetition, and perplexity under the \texttt{facebook/opt-1.3b} \cite{zhang-2022}, defined in Appendix \ref{app:analysis}. 
        We plot each models' error rates against each of these attributes (Figure \ref{fig:analysis}) with the goal of identifying trends in these attributes that may contribute to disparity in model performance. 
        
        Model behavior patterns in different ways for each attribute. For word count, models exhibit highly variable performance on low-word count texts, but typically improve as the text length increases. However, certain of the DeTeCtive variants (TuringBench, OUTFOX) again struggle with longer texts, particularly as texts become longer than 400 words. Model performance is not strongly tied to punctuation, except in cases where it is very rare ($<1\%$): in this case, performance varies greatly, though this reflects only 0.7\% of samples.

        Perhaps most informative is repetition: model families exhibit substantially different error patterns based on repetition. DeTeCtive and FDG models show minimal error with low repetition, with more frequent errors on more repetitive texts, while most others exhibit the opposite behavior. Perplexity is also quite informative: most models achieve strong performance on low-perplexity texts, excepting DeTeCtive (OUTFOX). Some models then degrade substantially when perplexity increases (RADAR, BiScope (Arxiv), DeTeCtive (TuringBench/OUTFOX)), while the others maintain stable performance.

        These observations demonstrate that performance variations can be accounted for by some of these attributes: systems must accommodate a wide variety of possible lengths, repetitions, and perplexities. Models efficacy varies substantially across these attributes, and resulting implementations should account for these textual characteristics and how they may impact performance.

\section{Conclusions}
  In this work we demonstrate that the choice of datasets and metrics critically influences our assessment of machine-generated text detection system capabilities. Model performance varies substantially across different evaluation frameworks, with each system under specific metric and data conditions. This underscores the necessity of context-aware deployment: no single solution performs optimally across all use cases.

  Motivated by our analysis, we recommend system description include: (a) a discussion of the impact of dataset composition and class imbalance on reported performance, (b) rigorous definitions and motivations for all metrics used, and (c) a discussion of potential harms from metric-task misalignment based on intended use-cases (e.g., false positives in academic settings). 
  
  We also suggest practitioners use datasets that are diverse as possible in terms of length, punctuation, repetition, and perplexity, as well as more general attributes like domain, style, and  authorship. As different models excel at different types of data, more diverse evaluation data is required for accurate comparison. With regard to metrics, practitioners should focus on metrics that cover intended uses, again using different metrics that accurately cover system strengths and weaknesses. These metrics should be carefully considered and motivated particularly with regard to class imbalance in the data, and practitioners should discuss the impact of imbalance on their results along with reported metrics.

  For both datasets and metrics, a standardized set of benchmarks would be a useful tool for those developing models. For datasets, we've seen attempts in this direction (eg. \cite{dugan2024,wang2024}, etc), but these benchmarks are static, and thus won't reflect the rapid pace of advancement in generative models. For metrics, we find that most work eschews any discussion or motivation of metrics and this needs to be addressed. Depending on the intended application of a system and the datasets involved, the relevant metrics may be different, and thus comparison becomes difficult. Overall, it is key that both datasets and metrics are diverse and clearly motivated.

  While current systems demonstrate tremendous potential, their practical utility hinges on addressing these complexities.  Machine-generated text detection remains challenging, with the continuous emergence of new generative models and adversarial attacks demanding corresponding advances in detection methods. The effective and fair deployment of these methods on diverse, real-world scenarios depends on rigorous assessment, and this requires crucial awareness of the impacts of datasets and metrics.

\section{Ethical Considerations}
The primary ethical consideration surrounding this work is the ethical application of machine-generated text detection models. We aim to avoid making claims about the general usage of these models, and whether it is appropriate, but understand there are ethical implications for employing automated systems for decision making that may negatively impact stakeholders. Our work focuses on potential pitfalls in the evaluation of machine-generated text detection systems. These systems are currently in use in areas such as automated grading where they have the potential for substantial harms, particularly for false positives, in which a student may be punished unfairly. To this end our work is vital: better understanding of the risks of these models is necessary in order to make informed decisions about the deployment of such systems. 

\section{Limitations}
While our evaluation spans a diverse range of models, it is not exhaustive—many systems and benchmarks fall outside our analysis. We demonstrate that our core findings (evaluation variance on different datasets, importance of metrics, and relatively weak performance on human-written texts) hold across a representative sample of prevalent models available for practitioners, but these conclusions are necessarily bounded by the scope of our study. This limitation is particularly salient given the rapid evolution of both generative LLMs and detection methods: as new models and datasets emerge, the detection landscape will continue to shift.

We are similarly limited in language (we focus only on English) and datasets (we select a sample of available datasets). While we are optimistic that our results are likely to hold for other datasets and languages, further verification would be necessary.

Rather than providing a definitive assessment of all available systems, languages, and datasets, our work highlights the critical need for rigorous evaluation methodologies, a principle we validate through our selected models and datasets, and one we argue extends to future research in this evolving field.


\section{References}\label{sec:reference}

\bibliographystyle{lrec2026-natbib}
\bibliography{custom}

\appendix

\section{Metrics}
\label{app:metrics}
We here define the metrics used for evaluation. We use the \texttt{numpy} and \texttt{scikit-learn} packages to calculate these metric.

\begin{itemize}
    \item \textbf{Precision}: the ratio of true positive predictions (TP) to the total number of positive predictions (true and false): $\frac{TP}{TP+FP}$
    \item \textbf{Recall}: the ratio of true positive predictions to the total number of positive samples (true positives and false negatives)in the dataset: $\frac{TP}{TP+FN}$
    \item \textbf{F1 Score}: the harmonic mean mean of precision and recall: $\frac{2 * prec * rec}{prec + recall}$
    \item \textbf{Accuracy}: the ratio of correctly classified samples to the full dataset: $\frac{TP + TN}{TP+TN+FP+FN}$
    \item \textbf{Average Recall}: the mean of recall for the positive class and the negative class.
    \item \textbf{AUROC}: area under the receiver operator characteristic (ROC) curve. This curve is defined as 
    
    $(FPR(x), TPR(x)), x \in [-\infty, +\infty]$ 
    
    For each value of $x$, we plot the false positive and true positive rate; for most models $x$ is a probability such that $x \in [0, 1]$. An AUROC of 1 means the model is perfect; .5 means the model performs at chance.

    \item \textbf{TPR @ FPR 1\% / .01\%}: True positive rate at a specific false positive rate. This is defined as the best possible percentage of true positives such that the false positive rate is below the given threshold.
\end{itemize}

Precision, recall, F1 score, accuracy, and average recall are binary classification based metrics: we use the model labels where provided, and classify based on thresholds where not provided. We use a default threshold of 0.5. In Section \ref{sec:metrics}, we also experiment using the threshold that optimizes equal error rate (EER), calculated by finding the point where false positive rate (FPR) and false negative rate (FNR) are closest.
\section{System Descriptions}
\label{app:systems}

        \subsection{Binoculars \cite{hans2024}} 
        The Binoculars system uses two pre-trained LLMs. The system calculates the perplexity of the text in question using an \textit{observer} and a \textit{performer}, and computes a metric called cross-perplexity, which is a strong signal for machine-generated text. Binoculars requires two LLMs but no training, is free and open-source, and they report excellent performance on a variety of domains, as well as on shortened texts.

        We use the implementation provided at \url{https://github.com/ahans30/Binoculars}.

        \subsection{BiScope \cite{guo2024-1}} 
        The BiScope model is based on the assumption that losses between a token and the preceding as well as following tokens can be indicative of AI generated text. They propose a bi-directional method, and train a classifier based on these statistics. BiScope is trained on four different corpora: two "short" text types (Yelp and Arxiv), and two "longer" text types (Creative and Essay): we evaluate these four variants. We do not evaluate the Code subset, as this does not align with our testing scenario.

        We use the implementation provided at \url{https://github.com/MarkGHX/BiScope}: they do not provide an explicit "best" model for each domain, so we train each of our four variants using all the provided data from the respective domains.
        
        \subsection{Fast-DetectGPT \cite{bao2023}} 
        Fast-DetectGPT (\textsc{fdg}) is an upgrade over the previously established DetectGPT \cite{mitchell2023}. They calculate a \textit{conditional probability curvature} of a given text, following the assumption that perturbing a machine generated text will lead to a lower conditional probability. Like Binoculars, \textsc{fdg} requires two LLMs: a sampling model and a scoring model, and it shows strong performance on a variety of domains. We evaluate three settings using different models for scoring: \texttt{gpt-neo-2.7b} for speed, \texttt{gpt-j-6b} as a slower but more accurate model\footnote{Following their work, we use \texttt{gpt-j-6b} as the reference model and \texttt{gpt-neo-2.7b} as the scoring model. For more, see \newcite{bao2023}.}, and \texttt{falcon-7b} for maximal accuracy.

        We use the implementation provided at \url{https://github.com/baoguangsheng/fast-detect-gpt}.

        \subsection{RADAR \cite{hu2023}} 
        The RADAR system uses adversarial training of a detector and paraphraser. The paraphraser attempts to fool the detector, and joint learning allows their model to adapt to new paraphrasing techniques. 

        We use the implementation provided at {https://github.com/IBM/RADAR}.

        \subsection{DeTeCtivE \cite{guo2024}} 
        The DeTeCtive model relies on a contrastive learning framework, calculating distances between samples which allows for encoding distinctive features based on the author. These encodings are then used at inference time, and the text in question is compared to relevant vectors using K-Nearest Neighbor (KNN) classification. DeTeCtive has four relevant training sets: MAGE (Deepfake), M4, TuringBench and OUTFOX: we evaluate each of these variants.

        We use the implementation provided at \url{https://github.com/heyongxin233/DeTeCtive}.
        
        \subsection{Zippy \cite{thinkst2023}}
        Unlike the above, which rely on heavily on pre-trained LLMs to calculate metrics, Zippy is compression based, relying on a variety of compression ratios to indirectly measure perplexity of a given text. Zippy starts by 'seeding' a compression with machine-generated text, and then assesses the difference in compression from just this seed to the addition of the text in question. Zippy reports strong performance compared to LLM-based models, and is independent from any models, making it an enticing option. 

        We use the implementation provided at \url{https://github.com/thinkst/zippy}.

\subsection{Transformer models}
    We use the \texttt{huggingface} platform for model training and evaluation \cite{wolf2020}. We start with a brief hyperparameter tuning to optimize learning rate and batch size; we find that a learning rate of $2e-5$ and a batch size of eight is effective accross model types. We train each model for 10 epochs, with weight decay of 0.01, then keep the best performing model on the training data by F1 score for prediction. We note that this may bias models towards performance on this particular metric: in practice we find fine-tuning these models using F1 score as the target yields consistent performance across all metrics. 

\subsection{McGovern Model}
    We follow the method described by \newcite{mcgovern2025}. We use the \texttt{scikit-learn} package \newcite{pedregosa2011} to do feature extraction, generating n-grams ($n \in {2,3,4}$) for words, part-of-speech tags, and characters. We use \texttt{nltk}'s word and part-of-speech taggers \cite{bird2009}. For model training, we use the GradientBoostingClassifier provided by \texttt{scikit-learn}. We perform a grid search over estimators, learning rate, and depth, selecting 100 estimators, a learning rate of 0.01, and a max depth of 3.

\subsection{Stylo Model}
\label{app:features}
Features for the \textsc{stylo} model are given below, along with additional notes for clarity. The text was processed using SpaCy \cite{honnibal2020}; unless otherwise specified, all syntactic and semantic information was extracted from the SpaCy parse. We calculate each feature as a raw count and also an average based on sentence length. We use univariate feature selection to select the 100 best features for model training. We then utilize ensemble classification via \texttt{scikit-learn}: we incorporate the \texttt{GaussianNB}, \texttt{AdaBoostClassifier}, \texttt{LGBMClassifier}, \texttt{CatBoostClassifier}, and \texttt{RandomForestClassifier} models into \texttt{scikit-learn}'s \texttt{VotingClassifier}. We use default model parameters, and run a paraeter search of weightings, which yields even weighting except for the \texttt{LGBMClassifier} which receives double weight.

\subsection{Architecture}
For model training, inference, and evaluation we use Amazon AWS EC2 instances. For CPU evaluation, we use an r5.2xlarge instances, with 64 GB of memory and 16 CPUs. For GPU testing we use the g6e.xlarge instance type. This instance type has an NVIDIA L40S Tensor Core GPU with 48 GB of GPU memory, allowing us to experiment with models that have larger GPU memory requirements (notably Binoculars and the FDG systems require significant GPU memory).
    \begin{figure*}[h!]
\centerline{\includegraphics[width=\textwidth]{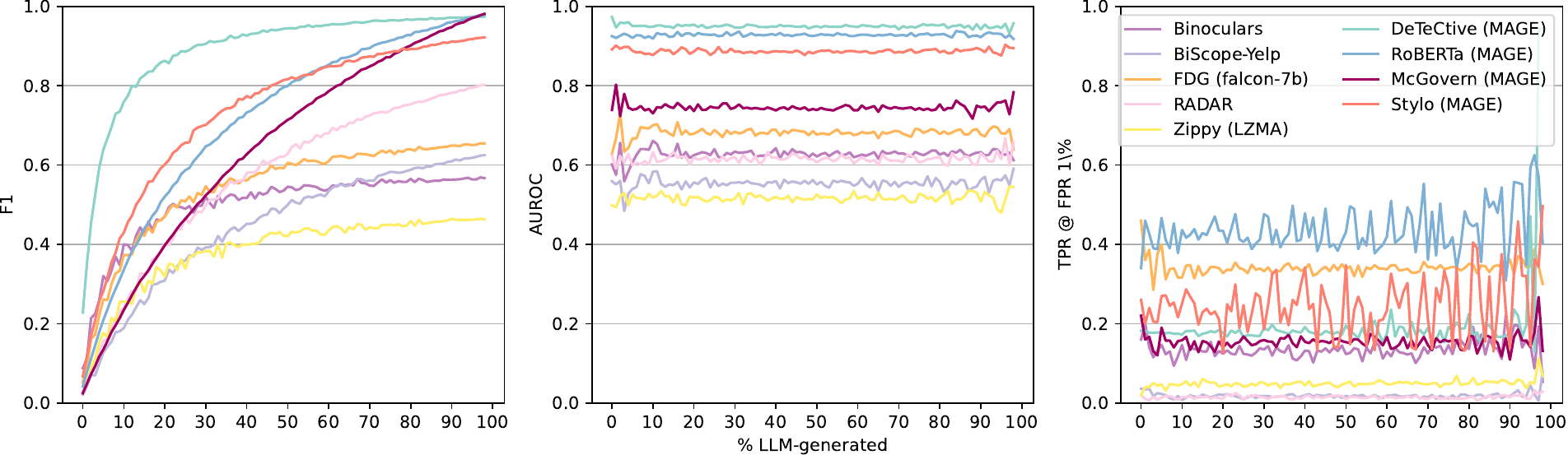}}
\caption{F1, AUROC, and TPR@FPR 1\% performance for imbalanced datasets. The x-axis reflects the percentage of samples from the positive (LLM-generated) class.}
\label{fig:balance} 
\end{figure*}

\section{Scores on imbalanced datasets}
\label{app:balance}
    Our test datasets are built to be balanced, but most systems described are evaluated on custom or previously used datasets that skew heavily towards LLM-generated classes. To investigate this bias, we systematically introduce artificial class imbalance into the \mage{} dataset. We select a sample of 5k items with a specific percentage $n$ being from the LLM-generated class. We evaluate these sample from $n=1$ where 1\% of the data is from the LLM-generated class to $n=100$ where the entire dataset is from the LLM-generated class. Figure \ref{fig:balance} compares F1, AUROC, and TPR@FPR 1\% metrics using this artificial imbalancing.

    Across all models, F1 scores exhibit logarithmic growth as the proportion of positive (LLM-generated) samples increases, while AUROC and TPR@FPR metrics remain stable regardless of class imbalance. This is due to the nature of the metrics: F1 is a threshold based metric and is inherently dependent on the number of true positives, false positives, and false negatives. With very few true samples, the score is dominated by false positives, while with very few false samples, recall can be inflated by classifying every sample as 1. AUROC and TPR metrics apply at all thresholds and are not affected by class imbalance: the key factors are ratios, and thus scaling the number of samples doesn't affect their values.
    
    This helps explain the strong performance reported in system descriptions evaluated with F1 score, as the datasets evaluated tend to skew heavily towards LLM-generated texts (the original MAGE training set is 70.8\% LLM-generated; the RAID training set is 97.1\% LLM-generated). These data imbalances are motivated by the desire to capture a wide variety of LLM-based generations: datasets are created with many LLM outputs generated from a single human text. 
    
    TPR@FPR shows significantly higher variance than AUROC when the positive class dominates, suggesting reduced reliability for skewed datasets. This instability cautions against over-reliance on TPR@FPR in LLM-heavy evaluation scenarios.
\section{Attributes Used for Analysis}
\label{app:analysis}

We evaluate four attributes as potential contributors to model performance. These are defined as follows:

\begin{itemize}
    \item \textbf{Length}: we simply use the number of words of the text, defined by splitting the text on whitespace.
    \item \textbf{Punctuation}: the percentage of the text that is punctuation, defined using Python's \texttt{string.punctuation}.
    \item \textbf{Repetition}: the percentage of word types in the text that occur more than once.
    \item \textbf{Perplexity}: defined as the exponentiation of the average log-likelihood under a language model for each text ($PPL(x) = e^{-\frac{1}{N}\sum_{i=1}^{N} \log P(x_i | x_{<i})}$). We use the \texttt{facebook/opt-1.3b} model to calculate log likelihoods \cite{zhang-2022}.
\end{itemize}

\end{document}